\documentclass[letterpaper, 10 pt, conference]{ieeeconf}  

\IEEEoverridecommandlockouts                              

\newcommand{\roa}{\texttt{RoA}}

\usepackage{amsfonts}
\usepackage{graphicx}
\usepackage{algorithm2e}
\usepackage{amssymb}
\usepackage{amsmath}
\usepackage{comment}
\usepackage[noadjust]{cite}

\usepackage{wrapfig}
\usepackage{overpic}
\usepackage[cmtip,arrow]{xy}
\usepackage{pb-diagram,pb-xy}
\usepackage{tikz}

\usepackage{color}
\usepackage{colortbl}
\usepackage{multirow}

\usepackage{graphicx}
\usepackage{optidef}
\usepackage{ulem}
\usepackage{makecell}

\usepackage{float}
\usepackage{tabularx}
\usepackage{url}
\usepackage{hyperref}

\newenvironment{myitem}
{
  \vspace{-0.04in}
    \begin{list}{$\circ$ }{}
        \setlength{\topsep}{0pt}
        \setlength{\parskip}{0pt}
        \setlength{\partopsep}{0pt}
        \setlength{\parsep}{0pt}         
        \setlength{\itemsep}{0pt} 
	\setlength{\leftskip}{-12pt}
}
{
    \end{list} 
    \vspace{-0.04in}
}

\newcommand{\cD}{\mathcal{D}}

\newcommand{\cF}{{\mathcal F}}

\newcommand{\cX}{{\mathcal X}}

\newcommand{\rM}{{\mathrm M}}

\newcommand{\sCG}{\mathsf{CG}}

    \newfont{\mbf}{msbm10 scaled 1100}

\newcommand{\mvmap}{\rightrightarrows}

\newcommand{\sMG}{{\mathsf{ MG}}}

\def\corcommstyle{\bf\small\tt}

\def\corrl #1<<#2||#3>>{
\if\visiblecomments y
  \begin{quote} {\corcommstyle $<<$COMMENT$>>$ {\color{red}#1\marginpar{!!}}\\$<<$OLD$<<$} \end{quote}

{\color{red} 
 #2
 }

  \begin{quote} {\corcommstyle ==NEW== } \end{quote}
   \noindent\hrulefill
 
\vspace{-10pt} 
 
 \noindent\hrulefill
 
 \vspace{-10pt} 
 
 \noindent\dotfill
 
  #3
  
   \noindent\dotfill 

\vspace{-10pt} 
 
 \noindent\hrulefill
 
 \vspace{-10pt} 
 
 \noindent\hrulefill
  \begin{quote} {\corcommstyle $>>$END$>>$ } \end{quote}
 \else
  #3
 \fi
}

\long\def\longcorrl #1<<#2||#3>>{
\if\visiblecomments y
  \begin{quote} {\corcommstyle $<<$COMMENT$>>$ {\color{red}#1\marginpar{!!}}\\$<<$OLD$<<$} \end{quote}
 
 {\color{red}

  #2
  
  }
  
  \begin{quote} {\corcommstyle ==NEW== } \end{quote}
  
    \noindent\hrulefill
 
\vspace{-10pt} 
 
 \noindent\hrulefill
 
 \vspace{-10pt} 
 
 \noindent\dotfill
 
  #3
  
   \noindent\dotfill 

\vspace{-10pt} 
 
 \noindent\hrulefill
 
 \vspace{-10pt} 
 
 \noindent\hrulefill
  \begin{quote} {\corcommstyle $>>$END$>>$ } \end{quote}
 \else
  #3
 \fi
}

\def\mlabel #1
{
  \if\visiblecomments y
     \marginpar[\flushright \bf \footnotesize #1]{\bf \footnotesize #1}
  \fi
}

\def\flabel #1
{
  \if\visiblecomments y
       \hbox{\bf\footnotesize #1}
  \fi
}

\def\corrq #1<<#2>>{
\if\visiblecomments y
  \begin{quote} {\corcommstyle $<<$COMMENT$>>$ {\color{red}#1}\marginpar{!!}\\$<<$BEG$<<$} \end{quote}
  \noindent\hrulefill
 
\vspace{-10pt} 
 
 \noindent\hrulefill
 
 \vspace{-10pt} 
 
 \noindent\dotfill

  #2
 
  \noindent\dotfill 

\vspace{-10pt} 
 
 \noindent\hrulefill
 
 \vspace{-10pt} 
 
 \noindent\hrulefill 
  \begin{quote} {\corcommstyle $>>$END$>>$ } \end{quote}
 \else
  #2
 \fi
}

\long\def\longcorrq #1<<#2>>{
\if\visiblecomments y
  \begin{quote} {\corcommstyle $<<$COMMENT$>>$ #1\marginpar{!!}\\$<<$BEG$<<$} \end{quote}
  \noindent\hrulefill
 
\vspace{-10pt} 
 
 \noindent\hrulefill
 
 \vspace{-10pt} 
 
 \noindent\dotfill

  #2

  \noindent\dotfill 

\vspace{-10pt} 
 
 \noindent\hrulefill
 
 \vspace{-10pt} 
 
 \noindent\hrulefill 
  \begin{quote} {\corcommstyle $>>$END$>>$ } \end{quote}
 \else
  #2
 \fi
}

\def\corrc #1<<>>{
\if\visiblecomments y
  \begin{quote} {\corcommstyle $<<$COMMENT$>>$ \color{red} #1\marginpar{!!}} \end{quote}
\fi
}

\def\corre #1<<#2||#3>>{
\if\visiblecomments y
  #3\marginpar{\corcommstyle #1}
 \else
  #3
 \fi
}

\long\def\longcorre #1<<#2||#3>>{
\if\visiblecomments y
  #3\marginpar{\corcommstyle #1}
 \else
  #3
 \fi
}

\def\corrs #1<<#2||#3>>{
\if\visiblecomments y
  #3\marginpar{\corcommstyle #2 $\rightarrow$ #3\\ #1}
 \else
  #3
 \fi
}

\def\corro #1<<#2||#3>>{
#2}

\def\corrn #1<<#2||#3>>{
#3}

\long\def\longcorro #1<<#2||#3>>{
#2}

\long\def\longcorrn #1<<#2||#3>>{
#3}

\long\def\underconstruction #1<<<#2>>>{
\if\visiblecomments y
  \begin{quote} {\corcommstyle $<<$UNDER CONSTRUCTION - BEGIN$>>$ #1\marginpar{!!}} \end{quote}
  #2
  \begin{quote} {\corcommstyle $>>$UNDER CONSTRUCTION - END$>>$ } \end{quote}
 \else
 \fi
}

\def\showcomments{
  \let\visiblecomments y
}

\def\hidecomments{
  \let\visiblecomments n
}

\showcomments

\title{\LARGE \bf Data-Efficient Characterization of the Global Dynamics\\ of Robot Controllers with Confidence Guarantees}

\author{Ewerton R. Vieira$^{3,5}$, Aravind Sivaramakrishnan$^{1}$, Yao Song$^{6}$, Edgar Granados$^{1}$,\\ Marcio Gameiro$^{2}$, Konstantin Mischaikow$^{2}$, Ying Hung$^{6}$, and Kostas E. Bekris$^{1}$
\thanks{$^{1}$ Dept. of Computer Science, Rutgers, NJ, USA. $^{2}$ Dept. of Mathematics, Rutgers, NJ, USA. $^{3}$ DIMACS, Rutgers, NJ, USA. $^{4}$ ICMC, Universidade de S\~{a}o Paulo, S\~{a}o Carlos, S\~{a}o Paulo, Brazil. $^{5}$ IME, Universidade Federal de Goi\'{a}s,
Goi\^{a}nia, GO, Brazil. $^{6}$ Dept. of Statistics, Rutgers University, NJ, USA. E-mail: {\tt \{er691,kb572\}@rutgers.edu}.}%
\thanks{This work is supported in part by NSF HDR TRIPODS award 1934924. MG and KM were partially supported by NSF under awards DMS-1839294, DARPA HR0011-16-2-0033, and NIH R01 GM126555. MG was partially supported by CNPq grant 309073/2019-7.}
}

\begin{document}
\maketitle
\thispagestyle{empty}
\pagestyle{empty}
\begin{abstract}
    This paper proposes an integration of surrogate modeling and  topology to significantly reduce the amount of data required to describe the underlying global dynamics of robot controllers, including closed-box ones. A Gaussian Process (GP), trained with randomized short trajectories over the state-space, acts as a surrogate model for the underlying dynamical system. Then, a combinatorial representation is built and used to describe the dynamics in the form of a directed acyclic graph, known as {\it Morse graph}. The Morse graph is able to describe the  system's attractors and their corresponding regions of attraction (\roa). Furthermore, a pointwise confidence level of the global dynamics estimation over the entire state space is provided. In contrast to alternatives, the framework does not require estimation of Lyapunov functions, alleviating the need for high prediction accuracy of the GP. The framework is suitable for data-driven controllers that do not expose an analytical model as long as Lipschitz-continuity is satisfied.  The method is compared against established analytical and recent machine learning alternatives for estimating \roa s, outperforming them in data efficiency without sacrificing accuracy. Link to code: \url{https://go.rutgers.edu/49hy35en}

\end{abstract}


\section{Introduction}
\label{sec:intro}

Multiple tools have been developed to estimate the region of attraction (\roa) of a dynamical system \cite{giesl2015review,vannelli1985maximal,bansal2017hamilton}. These tools are useful for understanding the conditions under which a controller can be safely applied to solve a task. Finding the \textit{true} \roa \ of a controlled system is challenging. Thus, many efforts try to estimate the largest possible set contained in the true \roa. For closed-box systems, such as learned controllers that do not provide an analytical expression, it is impractical to apply Lyapunov methods directly. Many non-Lyapunov methods often have significant data requirements so as to estimate \roa s effectively.  

\begin{figure}
    \centering
    \includegraphics[width=.99\columnwidth]{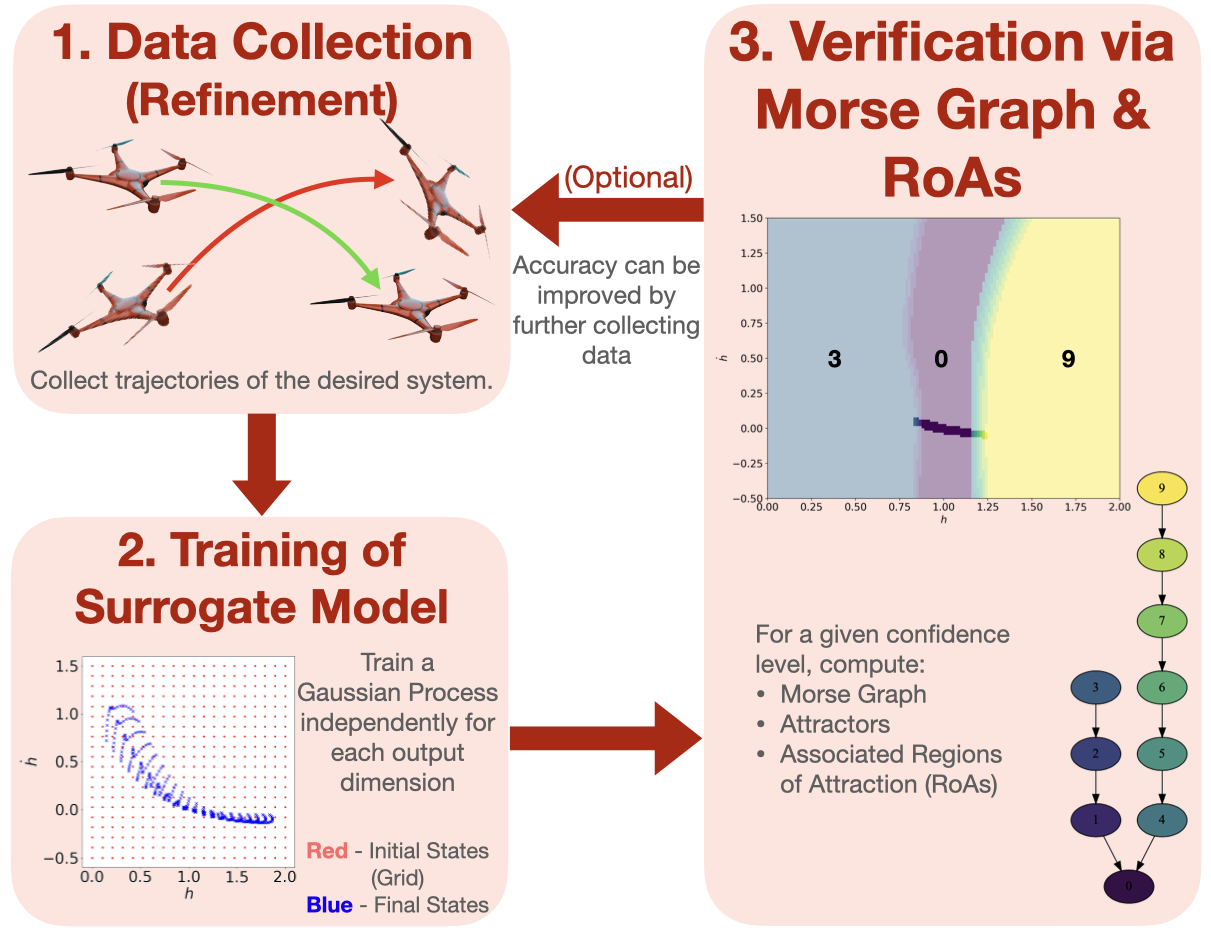}
    \vspace{-.2in}
    \caption{Overview of the proposed framework: 1) initial data collection; 2) a Gaussian Process (GP) is trained as a surrogate model; 3) computation of Morse Graph and the Region of Attraction (\roa) for verification. For an optional refinement, steps 1-3 can be repeated as necessary.}
    \label{fig:overview}
    \vspace{-0.2in}
\end{figure}

This paper addresses the problem of finding \roa s of controllers with unknown dynamics by proposing an efficient way to use data. It explores surrogate modeling together with topological tools not only to identify the \roa \ for a specific goal region but also to  describe the global dynamics. This also includes data-driven controllers, where a key challenge in their application is \textit{verification}, i.e. explaining when the controller works and when it fails.  To achieve this objective, this work uses Gaussian Processes (GPs) as surrogate models to compute a {\it Morse graph}, which constructs a finite, combinatorial representation of the state space given access to a discrete-time representation of the dynamics. It achieves data efficiency and improved accuracy relative to alternatives that are either analytical tools (and can only be used for analytical systems) or learning-based frameworks. Fig. \ref{fig:overview} highlights the iterative nature of the approach.

In particular, the key contribution of this work is the use of GPs as a statistical surrogate model of the underlying controlled system, alongside the Morse Graphs framework to compactly describe the global dynamics. The integration results in data efficiency: significantly fewer samples of the underlying dynamics are necessary for an informative representation of the global dynamics.  Data efficiency in surrogate modeling is achieved by leveraging the effectiveness of Morse Graphs, alleviating the high prediction accuracy requirements typically required for this purpose. 

Furthermore, this integration allows working with trajectories that may not uniformly cover the state space of the underlying system. Prior efforts with topological tools and combinatorial decompositions of the underlying state space required sampling the dynamics uniformly over a grid-based discretization of the state space. A GP allows an incremental approach where the collection of additional data points is guided to minimize the uncertainty about the global dynamics. Additionally, GPs provide confidence levels on the accuracy of the results at each subset of the state space. 

    


\section{Related Work}
\label{sec:related}

Numerical methods that estimate the \roa \ given a closed-form expression of the system dynamics  include maximal Lyapunov functions (LFs)  and linear matrix inequalities (LMIs).
Ellipsoidal \roa \ approximation via LMIs \cite{pesterev2017attraction,pesterev2019estimation} has been used for mobile robots \cite{rapoport2008estimation, pandita2009reachability}, and LMI relaxations can also approximate the \roa \ of polynomial systems \cite{henrion2013convex}.
LFs constructed by restricting them to be sum-of-squares (SoS) polynomials \cite{parrilo2000structured} have been used in building randomized trees with LQR feedback \cite{tedrake2010lqr}, funnel libraries \cite{majumdar2017funnel} and stability certificates for rigid bodies \cite{posa2013lyapunov}. 

Reachability analysis \cite{bansal2017hamilton}, i.e., computing a backward reachable tube to obtain the \roa \ without shape imposition, for computing RoAs of dynamical walkers \cite{choi2022computation}, has been combined with machine learning to maintain safety over a given horizon \cite{gillulay2011guaranteed}. GPs can learn barrier functions for ensuring the safety of unknown dynamical systems \cite{akametalu2014reachability}. Similarly, barrier certificates (BCs) can identify areas for exploration to expand the safe set \cite{wang2018safe}.  

Machine learning can learn LFs by alternating between a learner and a verifier \cite{chen2021learning_hybrid, abate2021fossil}, or via stable data-driven Koopman operators \cite{mamakoukas2020learning}. Rectified Linear Unit (ReLU) activated neural networks can learn robust LFs for approximated dynamics \cite{chen2021}. The Lyapunov Neural Network \cite{richards2018lyapunov} can incrementally adapt the \roa's shape given an initial safe set. As an alternative, GPs can obtain a Lyapunov-like function \cite{lederer2019local}, or an LF can be synthesized to provide guarantee's on a controller's stability while training \cite{dai2021lyapunov}.

GPs are a popular choice to reduce data requirements while modeling dynamical systems \cite{wang2005gaussian}. Some of their applications in robotics include model-based policy search \cite{deisenroth2013gaussian}, modeling non-smooth dynamics of robots with contacts \cite{calandra2016manifold}, and stabilizing controllers for control-affine systems \cite{castaneda2021}. For \roa \ estimation problems, given an initial safe set computed using a Lyapunov function, a GP can approximate the model uncertainties on a discrete set of sampling points from the safe region while expanding it \cite{berkenkamp2016safe}. 

Topology has multiple applications in robotics, such as deformable manipulation and others \cite{bhattacharya2015topological,antonova2021sequential,ge2021enhancing, varava2017herding,pokorny2016high,carvalho2019long}. Morse theory can help incrementally build local minima trees for multi-robot planning \cite{orthey2020visualizing} and finds paths to cover 2D or 3D spaces \cite{acar2002morse}. In recent work \cite{morsegraph}, Morse graphs are shown to be effective in compactly describing the global dynamics of a control system without an analytical expression of its dynamics. To the best of the authors' knowledge, the current work is the first to apply  surrogate modeling with uncertainty quantification in conjunction with topological tools to identify the global dynamics of robot controllers.  





\section{Problem Setup}
\label{sec:problem}

This work aims to provide a data-efficient framework for the analysis of global dynamics of robot controllers based on combinatorial dynamics and order theory \cite{morsegraph, kalies:mischaikow:vandervorst:14,kalies:mischaikow:vandervorst:15,kalies:mischaikow:vandervorst:21}. Consider a non-linear, continuous-time control system:    \vspace{-.1in}
\begin{equation}\label{eq:dyn}
    \dot{x} = f(x,u),
    \vspace{-.05in}
\end{equation}
where $x(t) \in X \subseteq \mathbb{R}^M$  is the state at time $t$, $X$ is a compact set, $u: X \mapsto \mathbb{U} \subseteq \mathbb{R}^M$ is a Lipschitz-continuous control as defined by a deterministic control policy $u(x)$, and $f : X \times \mathbb{U} \mapsto \mathbb{R}^M$ is a Lipschitz-continuous function. Neither $f(\cdot)$ nor $u = u(x)$ are necessarily known analytically. For a given time $\tau >0$, let $\phi_\tau: X \rightarrow X$ denote the function  obtained by solving Eq. \eqref{eq:dyn} forward in time for duration $\tau$ from everywhere in $X$. A trajectory (or an \textit{orbit}) is defined as a sequence of states obtained by integrating Eq. \ref{eq:dyn} forward in time.

The analysis of the global dynamics can reveal the system's attractors, which include fixed points, such as a state that the control law manages to bring the system to; or limit cycles, such as a periodic behavior of the system. It will also reveal a Region of Attraction (\roa) which is a subset of the basin of attraction of an attractor $\mathcal{A}$. The basin of attraction is the largest set of points whose forward orbits converge to $\mathcal{A}$, or more formally, the maximal set $\mathcal{B}$ that has the property: 
\vspace{-.1in}
\[
\mathcal{A}=\omega(\mathcal{B}):= \bigcap_{n \in \mathbb{Z}^+} \mathrm{cl}\left(\bigcup_{k=n}^\infty \phi_\tau^k(\mathcal{B})\right)
\vspace{-.1in}
\]
where $\phi_\tau^k$ is the composition $\phi_\tau\circ \cdots \circ\phi_\tau$ ($k$ times) and $\mathrm{cl}$ is topological closure.

Since $f$ and $u$ are Lipschitz-continuous, $\phi_\tau$ is too; furthermore any \roa \  of Eq. \eqref{eq:dyn} is  an \roa \ under $\phi_\tau$. Hence, it is possible to study  Eq. \eqref{eq:dyn}  by analyzing the behavior of the dynamics according to $\phi_\tau$, which is not assumed, however, to be computable and available.


\begin{figure*}
    \centering
    \includegraphics[width=0.33\textwidth]{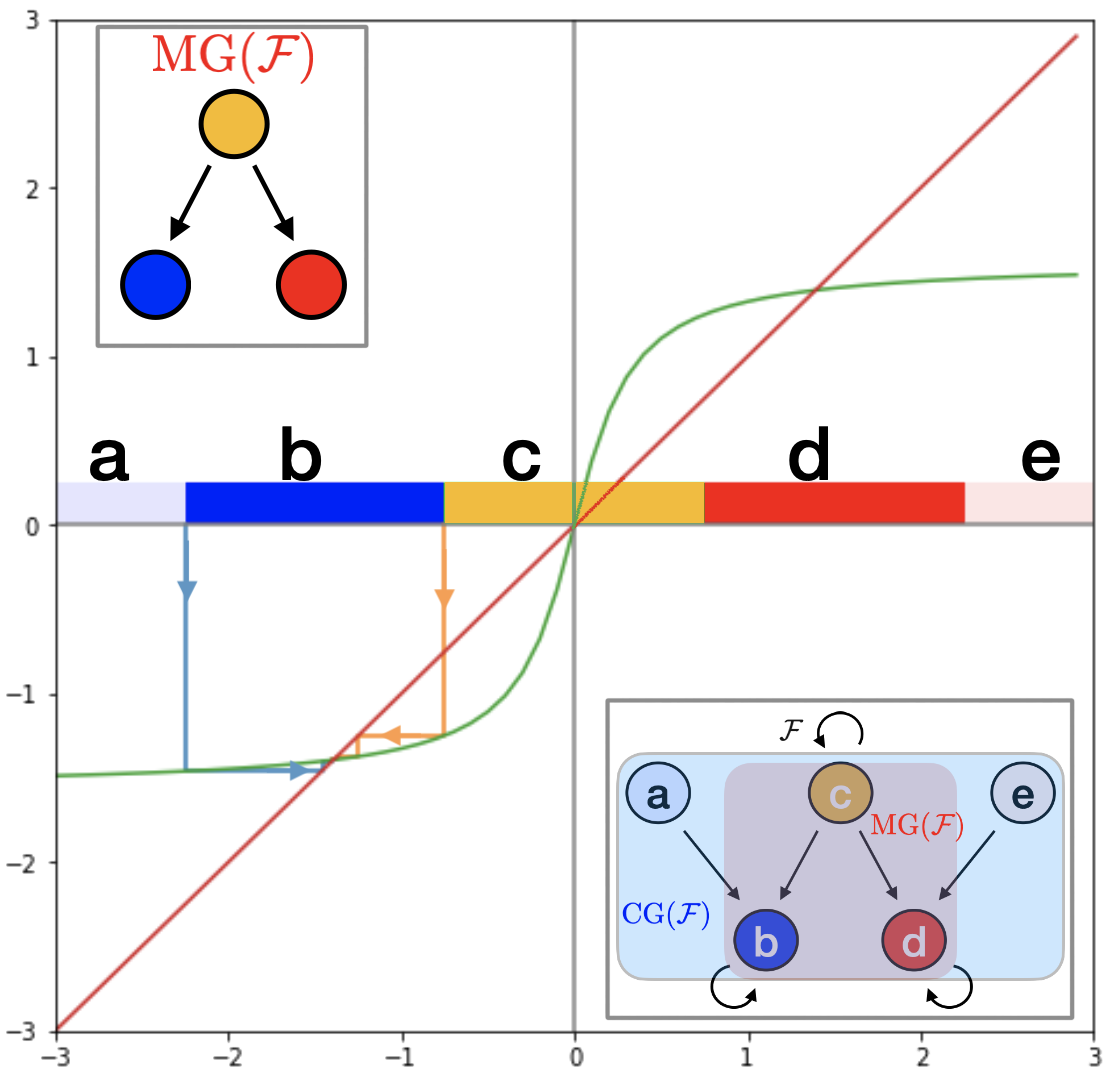}
    \includegraphics[width=0.65\textwidth]{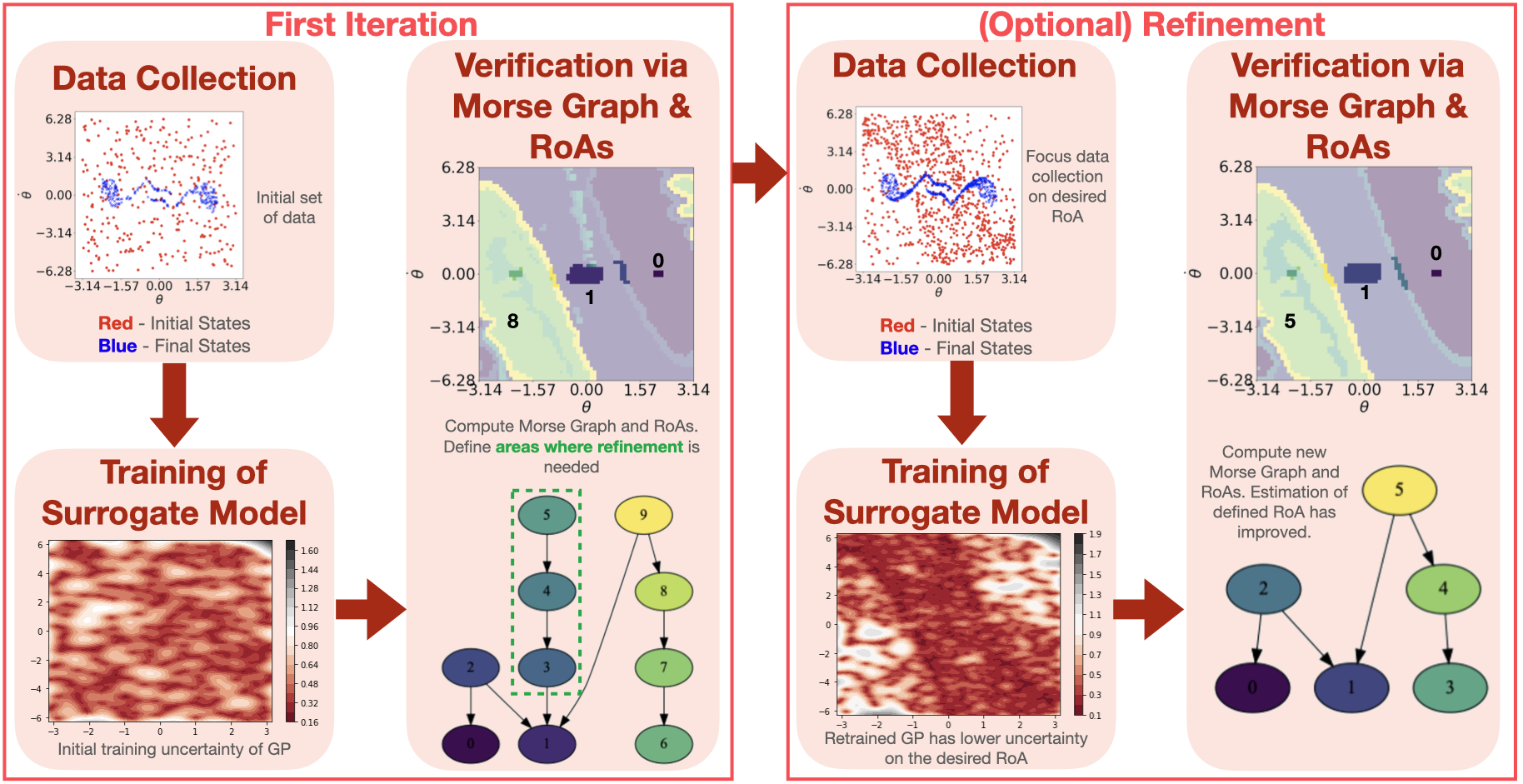}
    \vspace{-.15in} 
    \caption{
    \textbf{(Left)} 1D dynamics example $y=\arctan(x)$ is decomposed in cells [{\tt \bf{a}, \bf{e}}]. Forward propagation of {\tt \bf{b}} is depicted by arrows from its boundary. $\cF$ is a directed graph capturing reachable vertices (regions) from other vertices. Strongly connected components of $\cF$ result in $\mathrm{CG}(\cF)$. Finally, the Morse Graph $\mathrm{MG}(\mathcal{F})$ (nodes \{{\tt \bf b, c, d}\}) contains the attractors of interest. \textbf{(Center) }
    Proposed Method (initial step) for Pendulum (LQR). Data collected; heat map of initial GP; MG and \roa s of the initial GP. 
    \textbf{(Right)} Proposed method (refinement step) for Pendulum (LQR).
    Few new samples focused on the desired \roa \ are collected; retrained GP has lower uncertainty; MG and \roa s of the retrained GP.
    }
    \vspace{-.25in}
    \label{fig:overview_applied}
\end{figure*}

\section{Topological Framework and Uncertainty Quantification via GPs}
\label{sec:framework}

There are two key components for capturing meaningful conditions of the dynamics according to $\phi_\tau$. First, identifying  effective combinatorial representations of the attractors and  maximal \roa s. And second, to achieve data efficiency by employing GP-based surrogate modeling with uncertainty quantification. 

\textbf{Morse Graphs for Understanding Global Dynamics:} Fig. \ref{fig:overview_applied}(left) is used as a running 1-dim. example. The function $\phi_\tau$ is first approximated by decomposing the state space $X$ into a collection of regions $\cX$, for instance, by defining a {\it grid}. Fig. \ref{fig:overview_applied}(left) shows a grid on the interval $[-3,3]$ decomposed into sub-intervals $\bf{a}$ through $\bf{e}$. Given a region $\xi \in \cX$ (a cell), the system is forward propagated for multiple initial states within $\xi$ for a time $\tau$ to identify regions reachable from $\xi$. Consider, for example, the sub-interval $\bf{b}$ as such a cell $\xi$ in Fig. \ref{fig:overview_applied}(left). The arrows from the boundary of $\bf{b}$ depict the forward propagation of the dynamics where $\bf{b}$ maps to itself given the underlying dynamics

Then, a directed graph representation $\mathcal{F}$ stores each region in $\xi \in \cX$  as a vertex and edges pointing from $\xi$ to each region reachable from $\xi$. In Fig. \ref{fig:overview_applied}(left), $\mathcal{F}$ is the graph containing nodes given by the grid cells $\bf{a}$ to $\bf{e}$. Each edge represents a pair given by a cell and its image according to the dynamics. For instance, $(\bf{b}, \bf{b})$ and $(\bf{b}, \bf{c})$ are two edges added since $\bf{b}$ both maps to itself and also maps to $\bf{c}$. Condensing all the nodes belonging to a strongly connected components (SCCs) of $\mathcal{F}$ into a single node, results in the condensation graph $\mathrm{CG}\mathcal{(F)}$. Then, edges on $\mathrm{CG}\mathcal{(F)}$ reflect reachability according to a topological sorting of $\mathcal{F}$. In Fig. \ref{fig:overview_applied}(left), $\mathrm{CG}\mathcal{(F)}$ is the subgraph with nodes $\bf{a}$ to $\bf{e}$ and all non-self edges, i.e., $\mathrm{CG}\mathcal{(F)}$ has no cycles.

Since $\mathrm{CG}\mathcal{(F)}$ is a directed acyclic graph, it is also a partially ordered set (i.e., a poset). A \emph{recurrent set} is an SCC that contains at least one edge.  Finally, the \emph{Morse graph} of $\cF$, denoted by $\mathrm{MG}(\mathcal{F})$, is the subposet of recurrent set of $\mathrm{CG}\mathcal{(F)}$ (excluding single-node SCCs). In Fig. \ref{fig:overview_applied}(left), $\mathrm{MG}(\mathcal{F})$ is the graph with nodes $\bf{b}$, $\bf{c}$ and $\bf{d}$ and the corresponding edges between them. The Morse graph $\mathrm{MG}(\mathcal{F})$ captures the recurrent and non-recurrent dynamics by representing the recurrent sets of $\mathcal{F}$ as vertices and whose edges reflect reachability between these sets. The nodes of Morse graphs can contain attractors of interest. 


In summary, Morse graphs and \roa s are obtained by a four step procedure. {\bf 1)} State space decomposition and generation of input to represent $\phi_\tau$. {\bf 2)} Construction of the combinatorial representation $\cF$ of the dynamics given an outer approximation of $\phi_\tau$. {\bf 3)} Compute the \textit{Condensation Graph} $\sCG(\cF)$ and \textit{Morse Graph} $\sMG(\cF)$ by identifying recurrent sets/SCCs of $\cF$ and topological sort. {\bf 4)} Derive \roa s for the recurrent sets given the reachability of $\sCG(\cF)$. 


The \emph{state space decomposition} is an orthotope $X = \prod_{i=1}^n [a_i, b_i]$ (i.e., generalization of a rectangle for high-dim.), allowing for periodic boundary conditions. More specifically, a uniform discretization of $X$ is applied based on $2^{k_i}$ subdivisions in the $i$-th component resulting in a decomposition of the state space into $\prod_{i=1}^n 2^{k_i}$ cubes of dimension $n$. $\cX$ denotes the collection of these cubes. 

The \emph{input representation of $\phi_\tau$} is generated by the set of values of $\phi_\tau$ at the corner points of cubes in $\cX$. More precisely, let $V(\cX)$ denote the set of all corner points of cubes in $\cX$. The method computes the set of ordered pairs $\Phi_\tau(\cX):=\{(v,\phi_\tau(v)) \mid v \in V(\cX)\}$, by forward propagating the dynamics for time $\tau$
from all $V(\cX)$. Note that, no analytical version of $\phi_\tau$ is required, allowing a surrogate model to generate data $\Phi_\tau(\cX)$, as proposed in this work.

The \emph{combinatorial representation of the dynamics} is approximated by a \emph{combinatorial multivalued map} $\cF \colon \cX\mvmap \cX$, where vertices are $n$-cubes $\xi \in \cX$. The map $\cF$ contains directed edges $\xi \to \xi', \forall\ \xi, \xi' \in \mathcal{X}$ such that $\xi' \cap \Phi_\tau(\xi)\neq \emptyset$. The set of cubes identified by $\cF(\xi)$ are meant to capture the possible states of $\phi_\tau(\xi)$. Then, $\forall \xi \in \cX$ a multivalued map $\cF$ that satisfies:
\vspace{-.1in}
\begin{equation}
    \label{eq:OuterApproximation}
    \cF_{min}(\xi) := \{ \xi' \in \cX  \mid  \xi'\cap \phi_\tau(\xi) \neq \emptyset \} \subset \cF(\xi)
    \vspace{-.05in}
\end{equation}
is called an \emph{outer approximation} of $\phi_\tau$. Computation of $\cF_{min}$ is typically prohibitively expensive. But it is sufficient to find an outer approximation $\cF$, which still leads to mathematically rigorous results. The flexibility in defining an outer approximation provides versatility in its construction, which allows integration with a surrogate model.

\textbf{Surrogate Modeling and Uncertainty Quantification by GPs:} Assume that there is access to data of the form $\cD=\{ (x^n,y^n)\in X \times X \ |\ y^n=\phi_\tau(x^n) \text{ and } n=1, \ldots, N\}$, which may have Gaussian noise. In the pair $(x^n,y^n) \in \cD$, $x^n$ is an initial state of the system, and $y^n$ is the end state after forward propagating the dynamics (\ref{eq:dyn}) from $x^n$ for time $\tau$. In Figs ~\ref{fig:overview} and \ref{fig:overview_applied}(center and right), $(x^n,y^n)$ are denoted as red and blue points respectively. 
Let $\phi_{\tau, \ell}$ denote the $\ell$-th component of $\phi_\tau$, for $\ell=1, \ldots, M$ and assume that $\phi_{\tau, \ell}$ is the realization of $\phi_\tau$ from GP:
\vspace{-.1in}
\begin{equation}
\label{gp}
\phi_{\tau, \ell}(x)\sim GP(\beta_\ell, \sigma_\ell^2 k(x,x';\theta_\ell)), 
\end{equation}
where $\beta_\ell$ and $\sigma_\ell^2$ are the unknown mean and variance, and the correlation is defined by the kernel $ k(x,x';\theta_\ell)=
Corr(\phi_{\tau, \ell}(x), \phi_{\tau, \ell}(x');\theta_\ell)$  with $k(x,x;\theta_\ell)=1$, $k(x,x';\theta_\ell)=k(x',x;\theta_\ell)$ for $x,x'\in X$ and $\theta_\ell$ is a set of parameters associated with $k$.
 
The prediction for an untried $x\in X$ can be obtained by a $d$-dimensional multivariate normal distribution, $MN(\mu(x),\Sigma(x))$, where $\mu=(\mu_1,\cdots,\mu_d)$, $\mu_\ell(x)=E(\phi_{\tau, \ell}(x)|\cD)=\hat{\beta}_\ell+k(x;\hat{\theta}_\ell)^TK^{-1}(\hat{\theta}_\ell)(y_\ell^T-\hat{\beta}_\ell)$, and the covariance matrix $\Sigma(x)$ is a diagonal matrix with elements $\hat{\sigma}_\ell^2\left( 1-k(x;\hat{\theta}_\ell)^T K^{-1}(\hat{\theta}_\ell) k(x;\hat{\theta}_\ell)\right)$ assuming the $M$-dimensional outputs are independent, $\hat{\beta}_\ell$, $\hat{\sigma}_\ell$, and $\hat{\theta}_\ell$ are the maximum likelihood estimators, $k(x;\hat{\theta})=[k(x,x_n;\hat{\theta}), n=1,\cdots,N]$, and  $K(\hat{\theta})$ is an $N\times N$ matrix with elements $k(x_i,x_j;\hat{\theta})$ for $1\leq i,j \leq N$. 
\section{Proposed Integrated Solution}
\label{sec:proposed}

The proposed framework brings together topological tools for combinatorial dynamics and GPs. GPs are used as surrogate models to identify the global dynamics and \roa s of controllers, including data-driven ones without access to an analytical model. Fig. \ref{fig:overview_applied} (center and right) summarizes the method's application for a pendulum controlled by a linear quadratic regulator (LQR).  Overall, the method can be divided into the following steps: 

\begin{myitem}
    \item[1.] Collect data from the system.
    \item[2.] Apply GP regression to get an initial surrogate model with predictive mean $\mu$ and covariance function $\Sigma$.
    \item[3.] Compute the Morse Graph of the trained GP to obtain, with a given confidence level, the information about the global dynamics and the \roa s. 
    \item[4.] (Optional) To increase the confidence level, select state space points to collect more data as in step 1 to improve the accuracy of the representation and return to step 2.
\end{myitem}







\textbf{Step 1: Data Collection} Two procedures for data collection are explored. The first one collects short trajectories from random initial points in $X$ with a fixed duration of time. The second option collects time series data in the form of long trajectories, breaking them into smaller ones. Denote by $\cD=\{ (x^n,y^n) \in \mathbb{R}^M\times\mathbb{R}^M \ |\ y^n=\phi_\tau(x^n) \text{ for } n=1, \ldots, N\}$ the robot trajectory data collected.

\textbf{Step 2: GP regression} Given the training data $\mathcal{D}$ obtained in the previous step, a GP model is trained independently for each output dimension using a zero mean prior and a Mat\'{e}rn kernel with $\nu>1$. Learned controllers typically give Lipschitz-continuous functions, yet, not necessarily smooth. Hence kernels requiring less smoothness assumptions are ideal. As such, Mat\'{e}rn kernels provide better performance than a more common radial basis function kernel.

Let $\mu$ and $\Sigma$ denote the predictive mean and covariance functions of the GP, respectively. For a \emph{confidence level} $1-\delta$ and an $x\in \mathbb{R}^M$, since the GP model is trained independently for each output dimension, the \emph{confidence ellipsoid} is an $M$-dimensional hypercube $E^\delta_{\Sigma(x)}:=\prod_{n=1}^M I^{\alpha}_{n, \Sigma(x)}$, where $\alpha = 1-(1-\delta)^{1/M}$, $I^\alpha_{n, \Sigma(x)}=\left\{u\in \mathbb{R}\ |\ \| u-\mu_n(x)\| / \sigma(x)< z_{\alpha/2}\right\}$ is the confidence interval for the $n$-th output at $x$, and $z_{\alpha/2}$ is the corresponding critical value of the standard normal distribution. 

\begin{wrapfigure}{r}{0.5\columnwidth}
    \centering
    \vspace{-0.2in}
    \centering
        \includegraphics[width=0.5\columnwidth]{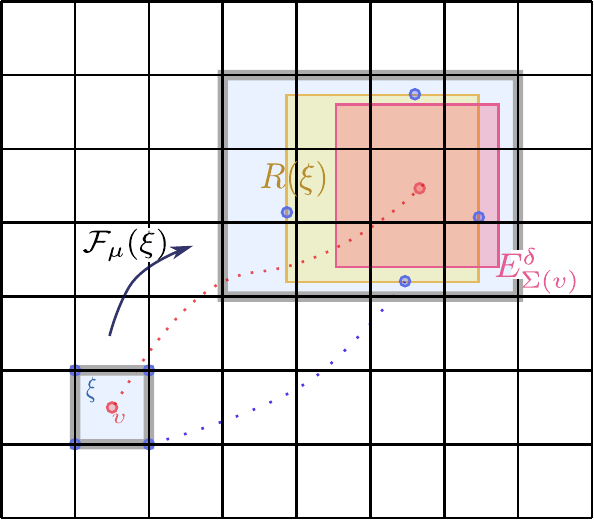}
    \vspace{-.25in}
    \caption{Pointwise confidence multi-valued map $\cF_\mu$.}
    \label{fig:pointwise}
    \vspace{-0.25in}
\end{wrapfigure}

\noindent \textbf{Step 3: Confidence level of Morse Graph and RoAs} Let $\mathcal{X}$ be a discretization of $X$ into cubes, $V(\mathcal{X})$ and $C(\cX)$ be the collections of corner (vertices) and the center points of the cubes in $\mathcal{X}$, respectively. From a trained GP with predictive mean $\mu$ and covariance functions $\Sigma$ generate the set $\rM(\cX):=\{(v,\mu(v)) \mid v \in V(\cX)\cup C(\cX)\}$ and let $E^\delta_{\Sigma(v)}$ be the confidence ellipsoid centered at $\mu(v)$ for $v\in C(\cX)$. Note the dependence of $E^\delta_{\Sigma(v)}$ on parameter $\delta$, which can be conveniently selected to maximize the confidence level to provide an accurate representation of the global dynamics.


For a given confidence level $1-\delta$, define a multivalued map based on the trained GP as follows:

\vspace{-.2in}
$$
\cF_\mu(\xi) := \left\{ \xi' \mid  \xi' \cap \left( E^\delta_{\Sigma(v)}\cup R(\xi) \right)\neq \emptyset\  \text{ for }\ v\in C(\xi)  \right\}, 
$$
where $R(\xi)$ is the smallest box containing $\{\mu(v) \ |\ v\in V(\xi)\}$ and $V(\xi)$ is the set of corner points (vertices) of $\xi \in \cX$, as in Fig. \ref{fig:pointwise}.   $\cF_\mu$ is refer as the {\it pointwise confidence multi-valued map}. Note that, for every $v \in C(\cX)$, the map $\cF_\mu$ contains the confidence ellipsoid centered at $\mu(v)$. Therefore, all cubes in $\cX$ have a pointwise confidence level of $1-\delta$ at their centers.

Finally, use the multivalued map $\cF_\mu$ to compute the Morse graph, the associated attractors and their \roa s. The CMGDB library \cite{CMGDB} and {RoA} implement topological computations of $\mathrm{MG}$ and \roa s. 
In fact, assuming that the unknown $\phi_\tau$ dynamics is a realization of the GP, the multivalued map $\cF_\mu$ captures any given realization of GP with a pointwise confidence level of 1-$\delta$ at the center points of the discretization.


\textbf{Step 4: Incremental Update}\label{subsec:update} After computing the \roa s and the attractors, the accuracy in estimating the \roa s can be further improved by iteratively collecting more data and performing steps 2 and 3 again. The new data can be randomly selected either in the whole state space or in the \roa\ of the interested attractor and it has to be consistent with the inital choice of the forward time $\tau$. The former results in a more accurate description of the global dynamics since it decreases the overall uncertainty. The latter focuses on increasing the accuracy of the desired \roa.

\textbf{Properties and Contribution:} The theoretical foundations for this line of work can be found in prior publications \cite{morsegraph, kalies:mischaikow:vandervorst:14,kalies:mischaikow:vandervorst:15,kalies:mischaikow:vandervorst:21}, where it is shown that a Morse Graph reflects the global dynamics of any continuous system under assumptions aligned with those of Section \ref{sec:problem}. When the system dynamics are generated by the (continuous) predicted mean $\mu$ of a GP, it is possible to incorporate the uncertainty estimate of the GP to obtain confidence levels on the global dynamics obtained by the Morse Graph \cite{bogdan2022gp}. Furthermore, under the assumption of sufficient data, $\mathcal{F}_{min}(\xi)\subset \mathcal{F}_\mu(\xi)$ for all $\xi\in\cX$. Thus, the global dynamics obtained via Morse Graph has a pointwise confidence level of $1-\delta$ for identifying the unknown dynamics. Relative to previous theoretical efforts, this work contributes: a) greater efficiency by using pointwise confidence guarantees as opposed to a global confidence level; b) an adaptive strategy for selecting samples guided by the Morse Graph and the GP; c) an effective solution for 2 to 4-dim. systems (relatively to 1-dim. examples in \cite{bogdan2022gp}); and d) implementation and experiments on models of robotic systems.


\textbf{Discussion on Computational Cost:} The cost of training the GP model is $O(M^4N^3)$, $N$ the size of the dataset and $M$ the dimension of $X$. Computing the Morse Graph and \roa s is $O(V+E + G^2)$ with $V$ the number of vertices in the directed graph $\cF_\mu$, $E$ the number of edges, and $G$ the size of the Morse graph (typically smaller than 32) \footnote{A detailed discussion of the computational costs can be found in \cite{Gramacy2020, morsegraph, Bush:Gameiro:Harker}.}. Thus, the total computational cost is $O((k+1)(M^4N^3 + V + E))$, with $k$ the number of incremental updates performed (Step 4). The memory requirement is $O(V + 2(MN)^2)$, where the predicted mean and variance of the GP model store two matrices with total entries $(MN)^2$ and the grid size is $V$.


\section{Experimental Evaluation}
\label{sec:experiments}

The proposed framework, {\tt GPMG}, is compared against alternatives from the literature for different dynamical systems and controllers (Table~\ref{table:systems_ctrls}). Section~\ref{subsec:quant-results} reports the following metrics for each benchmark: (a) Accuracy of \roa\ estimation, and (b) data efficiency, i.e., number of forward propagations of the true dynamics needed. Section~\ref{subsec:qual-results} describes the global dynamics and \roa \ discovered by {\tt GPMG}.

{\bf Systems and Controllers}: 
The \textbf{1D Quadrotor} (Quad) \cite{yuan2021safe} is stabilized at a given height, generating trajectories rolled out in the simulator. The {\bf Pendulum} (Pend) is governed by $m\ell^2 \Ddot{\theta} = mG \ell \sin{\theta} - \beta\theta + u$, given mass $m$, gravity $G$, pole length $l$, and friction coefficient $\beta$. The \textbf{Mountain Car} (Car) is the continuous version of the popular Reinforcement Learning (RL) benchmark \cite{moore1990efficient}. \textbf{Ackermann} (Ack) is a forward-only car-like first order system. \textbf{Lunar Lander} (Land) \cite{meditch1964problem} is governed by $\ddot{x} = - \frac{k \dot{m}}{m} - g$, with $k > 0$ the velocity of exhaust gasses, mass flow rate $\dot{m} \leq 0$, and $g = 1.62$ moon's gravity. The \textbf{Two-link Acrobot} (Acro) \cite{spong_acrobot} is controlled by a single torque between the links.

\begin{table}[H]
\centering
\vspace{-.1in}
\begin{tabular}{|c|c|c|c|c|}
\hline
\textbf{System} & \textbf{$X$} & $\mathbb{U}$ &   \textbf{Controllers}\\ \hline

{\scriptsize\texttt{1D-Quadrotor}\cite{yuan2021safe} } & ${\scriptstyle(z, \dot{z})}$ & $T$ &  {\scriptsize\texttt{Learned} }\\ \hline

{\scriptsize\texttt{Pendulum} } & ${\scriptstyle(\theta, \dot{\theta})}$ & $\tau$ &  {\scriptsize\texttt{Learned, LQR} }  \\ \hline

{\scriptsize\texttt{Mountain-Car} } & ${\scriptstyle(x, \dot{x})}$ & $\tau$  & {\scriptsize\texttt{Learned} }  \\ \hline

{\scriptsize\texttt{Ackermann \cite{corke2011robotics}} }& ${\scriptstyle(x,y,\theta)}$ & ${\scriptstyle(\gamma, V )}$ &   \makecell{{{\scriptsize\tt Learned, LQR}}, \\{{\scriptsize\tt Corke}}} \\ \hline

{\scriptsize\texttt{Lunar-Lander} } & ${\scriptstyle(h, \dot{h}, m)}$ & $\dot{m}$ &  {\scriptsize\texttt{TOC} }  \\ \hline

{\scriptsize\texttt{Acrobot \cite{spong_acrobot}} }  & ${\scriptstyle(\theta_1, \theta_2, \dot{\theta_1}, \dot{\theta_2})}$ & $\tau_2$ &  {\scriptsize\texttt{Hybrid, LQR}}  \\ \hline
\end{tabular}

\caption{\vspace{-.05in} Systems and controllers considered in the evaluation.}
\label{table:systems_ctrls}
\vspace{-.25in}
\end{table}

LQR linearizes the system to compute a gain $k$ used in the control law $u(x_t)=-k \cdot x_t$. A time-optimal controller (TOC)  for the lunar lander \cite{meditch1964problem} achieves soft landing by having a free-fall period and then switching to full-thrust until touchdown.
The learned controllers are Soft Actor-Critic policy networks \cite{Haarnoja2018SoftAO} trained to maximize the expected return $\mathbb{E}_{x_0 \sim X} [\sum_{t=0}^\tau \mathcal{R}(x_t)]$, where the reward function is $\mathcal{R}: X \rightarrow \{0,-1\}$. $\mathcal{R}(x_t) = 0$ iff  $x_t$ is within an $\epsilon$ distance from the goal state and $-1$ otherwise.
A hybrid controller takes two (analytical or learned) controllers $u_1(x), u_2(x), x\in X$, and applies one controller in predetermined subsets of the state space $X_1,X_2 \subset X$, i.e.: $\dot{x}_{t+1} = f(x_t,u_1)\ \text{if} \ x_t\ \in\ {X}_1,\ \text{and}\ f(x_t,u_2)\ \text{if}\ x_t \in {X}_2$.
\subsection{Quantitative Results}
\label{subsec:quant-results}


\textbf{Comparison Methods:} Two lyapunov-based analytical methods (\texttt{L-LQR} and \texttt{L-SoS}) are used as comparison (as in \cite{farsi2022piecewise,richards2018lyapunov}). Both use a linearized unconstrained form of the dynamics \cite{prajna2002introducing} to obtain a Lyapunov function (LF). \texttt{L-LQR} uses the solution of the Lyapunov equation $v_{LQR}(x)=x^TPx$ while \texttt{L-SoS} computes the LF as $v_{sos}(x)=m(x)^TQm(x)$ where $m(x)$ are monomials on $x$ and $Q$ is a positive semidefinite matrix. \texttt{L-SoS} is implemented with SOSTOOLS \cite{prajna2002introducing} and SeDuMi \cite{sturm1999using}. These methods cannot be used with data-driven controllers (like {\tt Learned}) since closed-form expression is required. The Lyapunov Neural Network ({\tt L-NN}) \cite{richards2018lyapunov} is a machine learning tool  for identifying RoAs. The Morse graph ({\tt TopMG}) employs the topological tools described but without any use of a surrogate model, instead it queries the true dynamics at every vertex of the discretization \cite{morsegraph}.

\noindent \textbf{RoA Estimations:} An approximation of the ground truth \roa \ for the goal is computed by considering a very high-resolution grid over $X$, and forward propagating for a long time horizon, or until the goal is reached. Table \ref{tab:tp} presents the ratio of the ground truth \roa\ volume identified by each method for all benchmarks. With the exception of the Land (TOC) benchmark, {\tt GPMG} consistently estimates a larger ratio of the \roa \ volume compared to alternatives.

\begin{table}[h]
    \centering
    \vspace{-.1in}
    \small
    \begin{tabular}{|c||c|c|c|c|c|}
        \hline
        \textbf{Benchmark} & {\tt L-NN} & {\tt L-LQR/SOS} & {\tt TopMG} & {\tt GPMG}    \\ \hline \hline
        Quad (Learned) & - & N.A. & - & \textbf{1.0} \\ \hline
        Pend (LQR) & \textbf{0.98} & \ 0.7 / 0.03  & 0.97 & 0.91 \\ \hline 
        Car (Learned) & - & N.A.& {\bf 1.0} & {\bf 1.0} \\ \hline
        Land (TOC) & - & N.A. & {\bf 1.0} & { 0.79} \\ \hline
        Ack (Learned) & 0.91 & N.A. & \textbf{1.0} & \textbf{1.0}\\ \hline 
        Acro (LQR) & 0.89 & 0.27 / 0.26 & 0.96 & \textbf{1.0} \\ \hline
        Acro (Hybrid) & 0.14 & N.A. & {0.99} & \textbf{1.0} \\ \hline
    \end{tabular}
    \caption{\vspace{-.05in} \roa \ ratios for the methods. Best values per row in bold.}
    \label{tab:tp}
    \vspace{-.25in}
\end{table}


Estimating larger ratios of the \roa , however, may also lead to False Positives (FP) -- incorrectly identifying a volume of the state space as being in \roa. The optional fourth step (Section \ref{subsec:update}) of the proposed framework is crucial to mitigate FPs. All other methods ({\tt L-NN, L-LQR, SOS} and {\tt TopMG}), which require access to the true dynamics model, have zero FP. {\tt GPMG} falsely labels $0\% - 2\%$ of $X$ (Pend (LQR)) and $1\% - 22\%$ of $X$ (Land (TOC)) as part of the \roa. These cases are further discussed in Section~\ref{subsec:qual-results}.

\begin{table}[h]
\centering 
\vspace{-0.1in}
 \begin{tabular}{|c||c|c|c||c|}
        \hline
        \textbf{Benchmark} & {\tt L-NN} & {\tt TopMG} & Ours: {\tt GPMG} & Dim   \\ \hline \hline
        Quad (Learned) & - & - & {\bf 25,000} & 2\\ \hline
        Pend (LQR) & 667.1M & {6.6M} & {\bf 120,000} & 2 \\ \hline 
        Car (Learned) & -  & {6.6M}  & {\bf 3,000} & 3 \\ \hline 
        Land (TOC) & - & 1M & {\bf 300,000} & 3 \\ \hline
        Ack (Learned) & 704.6M  & {520M} & {\bf 10,000} & 3 \\ \hline 
        Acro (LQR) & 5.7B  &  {1.1B} &  {\bf 100,000}& 4 \\ \hline
        Acro (Hybrid) & 533M & 2.1B &  {\bf 2.5M} & 4 \\ \hline
\end{tabular}
    \caption{\vspace{-.05in} Propagations required. Best values per row in bold.}
    \label{tab:steps}
\vspace{-0.25in}
\end{table}

\noindent \textbf{Data Efficiency:} The data efficiency of methods requiring access to the underlying dynamical system ({\tt L-NN, TopMG} and {\tt GPMG}) is measured using the total propagation steps required to estimate the \roa\ (Table ~\ref{tab:steps}).  \emph{The data requirements for {\tt GPMG} are $2 - 4$ orders of magnitude less than {\tt TopMG} and $3 - 5$ orders of magnitude less than {\tt L-NN}}. The learned controllers benefit the most from {\tt GPMG}, as it provides, in all cases, a good coverage of the \roa \ with significantly fewer propagations and without FP (false positives).

\subsection{Qualitative case studies}
\label{subsec:qual-results}
\begin{figure}
    \includegraphics[width=0.95\columnwidth]{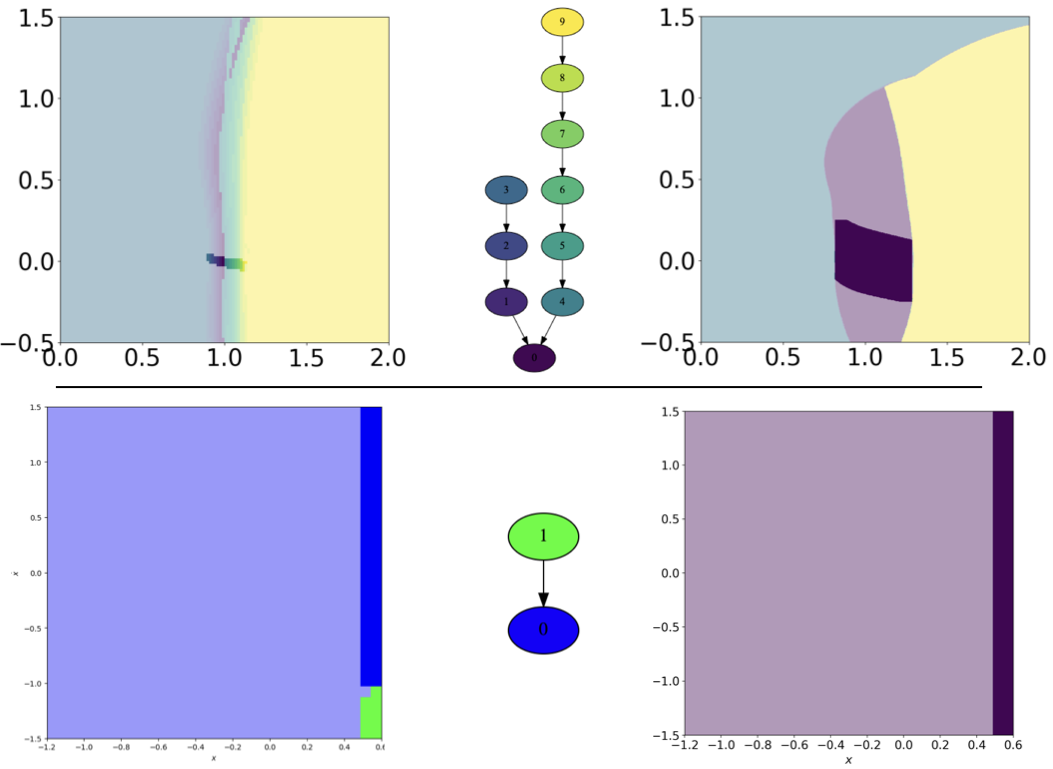}
    \vspace{-.2in}
    \caption{(Top) Left to right: Quad (Learned) \roa\ with confidence $12.5\%$, associated Morse Graph, and \roa\ with confidence $95\%$. (Bottom) Left to right: \roa\ of trained GP for Car, associated Morse graph, and \roa\ of the true dynamics.}
    \label{fig:1D_q}
    \vspace{-0.25in}
\end{figure}
 
\noindent \textbf{1D Quadrotor:} 100 random trajectories (average length 2.5s) that successfully reach the goal are used. To train the GP, trajectories are decomposed into short segments of $\tau = 0.3$s, giving a dataset $\cD=\{ (x_i,y_i) \in \mathbb{R}^M\times\mathbb{R}^M \ |\ y_i=\phi_\tau(x_i)\}_{i=1}^{800}$. For $12.5\%$ and $95\%$ confidence levels, {\tt GPMG} outputs an $\mathrm{MG}$ with 9 nodes.
The attractor discovered by {\tt GPMG} shows the whole state space divided into two regions represented by the left (nodes 1-3) and right (nodes 4-9) parts of the $\mathrm{MG}$. The trajectories of one region do not visit the other. These regions represent the system approaching the goal from above or below it. Node 0 (without leafs) of MG represents the region (Fig \ref{fig:1D_q}, darkest color) where all trajectories need to stabilize before reaching the goal region.

\begin{figure}
\centering
    \includegraphics[width=\columnwidth]{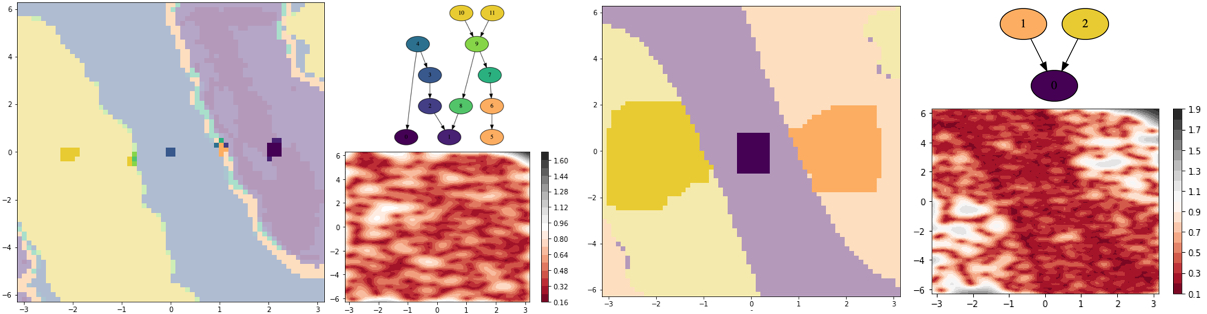}
\vspace{-.3in}
\caption{\roa s, Morse Set and Standard Deviation (left) of the initial surrogate model with $12.5\%$ confidence. After sampling $900$ more trajectories: \roa s, Morse Set and Standard Deviation (right) with $95\%$ confidence.}
    \label{fig:Pend_step90}
    \vspace{-0.28in}
\end{figure}

\textbf{Pendulum:} Initially, 300 random trajectories are used to train the initial surrogate model. The initial corresponding $\mathrm{MG}$, \roa s and the $\sigma$ are computed with $12.5\%$ confidence (Fig~\ref{fig:Pend_step90}, left). 
The initial procedure results in False Postives (FPs) that correspond to $2\%$ of $X$. To decrease the number of FPs, and to improve the accuracy of the estimated MG, Step 4 of the proposed framework is applied. 10 more samples are randomly selected at the boundary of and inside the \roa \ of the attractor of interest. When the procedure is repeated 90 times, a confidence level of $95\%$ is obtained, and all FPs are completely removed (Fig~\ref{fig:Pend_step90}, right).



\textbf{Mountain Car:} GP is trained with $300$ randomly sampled trajectories, each with duration $\tau=1s$, the predicted $\sigma$ is small enough to use a confidence level of $95\%$. The resulting $\mathrm{MG}$ (Fig~\ref{fig:1D_q} bottom) has two nodes and describes the expected global dynamics given by {\tt TopMG}.

\textbf{Lander:} {\tt GPMG} relies on the assumption that the underlying dynamical system $\phi_\tau$ can be realized by a GP that uses traditional kernels (Mat\'ern, Exponential, and Logistic for the sake of this discussion). The Lander benchmark violates this assumption since the goal region is a line $\{(0, 0, m) \ |\ m_1 \leq m\leq m_2 \}$ and not a system attractor. When trained with a Mat\'ern kernel, GPMG obtains FP. Even with additional data, the surrogate model does not satisfactorily capture the underlying dynamics of the system.

\textbf{Ackermann:} The GP is trained from 
$1000$ randomly sampled trajectories, each of duration $\tau=1s$. The predicted $\sigma$ of the GP is small enough to consider the $95\%$ confidence level. {\tt GPMG} outputs an $\mathrm{MG}$ with a single node representing the system's attractor, which exhibits periodic behavior, agreeing with the global dynamics captured by {\tt TopMG}. Hence, with significantly less data requirements, {\tt GPMG} successfully captures the global dynamics information (a torus-like shaped attractor). 
If longer trajectories ($\tau=40$s) are used, {\tt GPMG} outputs a $\mathrm{MG}$ with a single node representing the attractor ($0.008\%$ of $X$ by volume) without periodicity. This corresponds to the learned controller for Ackermann first performing periodic oscillations around the goal region before reaching it. 




\textbf{Acrobot:} For both the Acro (LQR) and Acro (Hybrid) benchmarks, the initial data collected used $1000$ random trajectories, each of duration $14$s (LQR) and $25$s (Hybrid). In both cases, the $\mathrm{MG}$ has a single node, and the identified attractor is a small set that contains the goal region. When additional data is provided, the uncertainty of {\tt GPMG} decreases, but the size of the attractor does not change notably, and the ratio of its volume of $X$ remains unchanged.



\section{Discussion}
\label{sec:discussion}


This work integrates surrogate modeling via GPs with topology tools, achieving a data-efficient framework for identifying the global dynamics (attractors and \roa s), even for closed-box systems. Tests on different benchmarks show the proposed method consistently identifying attractors with larger \roa\ coverage and a significant reduction in data requirements. A confidence level is also assigned to the global dynamics representation output. This novel approach allows the user to either work with a sparse dataset sacrificing confidence level associated with the Morse Graph; or guide the process for additional collection, increasing confidence levels for the Morse Graph by training GPs with low overall uncertainty. For dynamical systems that cannot be realized via a GP with a traditional kernel (e.g., Lander), non-conventional kernels can be explored to accommodate irregular input domains and non-Gaussian outputs.



On the theoretical side, the rate of convergence of the Morse Graph to the true dynamical system as a function of incremental samples is unknown. The decreasing rate of the overall standard deviation might provide insights to estimate this rate of convergence. Finally, the application to large and high-dimensional state spaces may still be challenging although it is shown to be more data efficient than alternatives. Possible strategies to mitigate this are non-uniform state space discretizations, such as adaptive schemes.




\clearpage


\bibliographystyle{IEEEtran}
\bibliography{IEEEabrv,example}

\begin{thebibliography}{10}
\providecommand{\url}[1]{#1}
\csname url@samestyle\endcsname
\providecommand{\newblock}{\relax}
\providecommand{\bibinfo}[2]{#2}
\providecommand{\BIBentrySTDinterwordspacing}{\spaceskip=0pt\relax}
\providecommand{\BIBentryALTinterwordstretchfactor}{4}
\providecommand{\BIBentryALTinterwordspacing}{\spaceskip=\fontdimen2\font plus
\BIBentryALTinterwordstretchfactor\fontdimen3\font minus
  \fontdimen4\font\relax}
\providecommand{\BIBforeignlanguage}[2]{{%
\expandafter\ifx\csname l@#1\endcsname\relax
\typeout{** WARNING: IEEEtran.bst: No hyphenation pattern has been}%
\typeout{** loaded for the language `#1'. Using the pattern for}%
\typeout{** the default language instead.}%
\else
\language=\csname l@#1\endcsname
\fi
#2}}
\providecommand{\BIBdecl}{\relax}
\BIBdecl

\bibitem{giesl2015review}
P.~Giesl and S.~Hafstein, ``{Review on computational methods for Lyapunov
  functions},'' \emph{Discrete \& Continuous Dynamical Systems-B}, vol.~20,
  no.~8, p. 2291, 2015.

\bibitem{vannelli1985maximal}
A.~Vannelli and M.~Vidyasagar, ``{Maximal Lyapunov functions and domains of
  attraction for autonomous nonlinear systems},'' \emph{Automatica}, vol.~21,
  no.~1, pp. 69--80, 1985.

\bibitem{bansal2017hamilton}
S.~Bansal, M.~Chen, S.~Herbert, and C.~J. Tomlin, ``{Hamilton-Jacobi
  reachability: A brief overview and recent advances},'' in \emph{CDC}, 2017.

\bibitem{pesterev2017attraction}
A.~V. Pesterev, ``Attraction domain estimate for single-input affine systems
  with constrained control,'' \emph{Automation and Remote Control}, vol.~78,
  no.~4, pp. 581--594, 2017.

\bibitem{pesterev2019estimation}
------, ``Attraction domain for affine systems with constrained vector control
  closed by linearized feedback,'' \emph{Autom. \& Remote Control}, vol.~80,
  no.~5, 2019.

\bibitem{rapoport2008estimation}
L.~B. Rapoport and Y.~V. Morozov, ``Estimation of attraction domains in wheeled
  robot control using absolute stability approach,'' \emph{IFAC}, vol.~41,
  no.~2, pp. 5903--5908, 2008.

\bibitem{pandita2009reachability}
R.~Pandita, A.~Chakraborty, P.~Seiler, and G.~Balas, ``{Reachability and RoA
  analysis applied to GTM dynamic flight envelope assessment},'' in \emph{AIAA
  CNC}, 2009.

\bibitem{henrion2013convex}
D.~Henrion and M.~Korda, ``{Convex computation of the RoA of polynomial control
  systems},'' \emph{IEEE Tran. on Automatic Control}, vol.~59, no.~2, pp.
  297--312, 2013.

\bibitem{parrilo2000structured}
P.~A. Parrilo, \emph{Structured semidefinite programs and semialgebraic
  geometry methods in robustness and optimization}.\hskip 1em plus 0.5em minus
  0.4em\relax California Institute of Technology, 2000.

\bibitem{tedrake2010lqr}
R.~Tedrake, I.~R. Manchester, M.~Tobenkin, and J.~W. Roberts, ``{LQR-trees:
  Feedback motion planning via sums-of-squares verification},'' \emph{IJRR},
  vol.~29, no.~8, 2010.

\bibitem{majumdar2017funnel}
A.~Majumdar and R.~Tedrake, ``Funnel libraries for real-time robust feedback
  motion planning,'' \emph{The International Journal of Robotics Research},
  vol.~36, no.~8, pp. 947--982, 2017.

\bibitem{posa2013lyapunov}
M.~Posa, M.~Tobenkin, and R.~Tedrake, ``Lyapunov analysis of rigid body systems
  with impacts and friction via sums-of-squares,'' in \emph{HSCC}, 2013, pp.
  63--72.

\bibitem{choi2022computation}
J.~J. Choi, A.~Agrawal, K.~Sreenath, C.~J. Tomlin, and S.~Bansal,
  ``{Computation of RoAs for Hybrid Limit Cycles Using Reachability},''
  \emph{arXiv:2201.08538}, 2022.

\bibitem{gillulay2011guaranteed}
J.~H. Gillulay and C.~J. Tomlin, ``Guaranteed safe online learning of a bounded
  system,'' in \emph{IROS}, 2011.

\bibitem{akametalu2014reachability}
A.~K. Akametalu, J.~F. Fisac, J.~H. Gillula, S.~Kaynama, M.~N. Zeilinger, and
  C.~J. Tomlin, ``Reachability-based safe learning with gaussian processes,''
  in \emph{CDC}, 2014.

\bibitem{wang2018safe}
L.~Wang, E.~A. Theodorou, and M.~Egerstedt, ``Safe learning of quadrotor
  dynamics using barrier certificates,'' in \emph{ICRA}, 2018.

\bibitem{chen2021learning_hybrid}
S.~Chen, M.~Fazlyab, M.~Morari, G.~J. Pappas, and V.~M. Preciado, ``{Learning
  lyapunov functions for hybrid systems},'' in \emph{HSCC}, 2021, pp. 1--11.

\bibitem{abate2021fossil}
A.~Abate, D.~Ahmed, A.~Edwards, M.~Giacobbe, and A.~Peruffo, ``{FOSSIL: a
  software tool for the formal synthesis of lyapunov functions and barrier
  certificates using neural networks},'' in \emph{HSCC}, 2021, pp. 1--11.

\bibitem{mamakoukas2020learning}
G.~Mamakoukas, I.~Abraham, and T.~D. Murphey, ``Learning stable models for
  prediction and control,'' \emph{IEEE Trans Robot}, 2020.

\bibitem{chen2021}
S.~Chen, M.~Fazlyab, M.~Morari, G.~J. Pappas, and V.~M. Preciado, ``Learning
  region of attraction for nonlinear systems,'' in \emph{2021 60th IEEE
  Conference on Decision and Control (CDC)}, 2021, pp. 6477--6484.

\bibitem{richards2018lyapunov}
S.~M. Richards, F.~Berkenkamp, and A.~Krause, ``{Lyapunov Neural Network:
  Adaptive stability certification for safe learning of dynamical systems},''
  in \emph{CoRL}, 2018.

\bibitem{lederer2019local}
A.~Lederer and S.~Hirche, ``{Local Asymptotic Stability Analysis and Region of
  Attraction Estimation with Gaussian Processes},'' in \emph{CDC}, 2019.

\bibitem{dai2021lyapunov}
H.~Dai, B.~Landry, L.~Yang, M.~Pavone, and R.~Tedrake, ``{Lyapunov-stable
  neural-network control},'' \emph{arXiv preprint arXiv:2109.14152}, 2021.

\bibitem{wang2005gaussian}
J.~Wang, A.~Hertzmann, and D.~J. Fleet, ``Gaussian process dynamical models,''
  \emph{Advances in neural information processing systems}, vol.~18, 2005.

\bibitem{deisenroth2013gaussian}
M.~P. Deisenroth, D.~Fox, and C.~E. Rasmussen, ``Gaussian processes for
  data-efficient learning in robotics and control,'' \emph{IEEE transactions on
  pattern analysis and machine intelligence}, vol.~37, no.~2, pp. 408--423,
  2013.

\bibitem{calandra2016manifold}
R.~Calandra, J.~Peters, C.~E. Rasmussen, and M.~P. Deisenroth, ``Manifold
  gaussian processes for regression,'' in \emph{2016 International Joint
  Conference on Neural Networks (IJCNN)}.\hskip 1em plus 0.5em minus
  0.4em\relax IEEE, 2016, pp. 3338--3345.

\bibitem{castaneda2021}
F.~Castañeda, J.~J. Choi, B.~Zhang, C.~J. Tomlin, and K.~Sreenath, ``Gaussian
  process-based min-norm stabilizing controller for control-affine systems with
  uncertain input effects and dynamics,'' in \emph{2021 American Control
  Conference (ACC)}, 2021, pp. 3683--3690.

\bibitem{berkenkamp2016safe}
F.~Berkenkamp, R.~Moriconi, A.~P. Schoellig, and A.~Krause, ``{Safe learning of
  RoAs for uncertain, nonlinear systems with Gaussian Processes},'' in
  \emph{CDC}, 2016.

\bibitem{bhattacharya2015topological}
S.~Bhattacharya, S.~Kim, H.~Heidarsson, G.~S. Sukhatme, and V.~Kumar, ``A
  topological approach to using cables to manipulate sets of objects,''
  \emph{IJRR}, vol.~34, no.~6, 2015.

\bibitem{antonova2021sequential}
R.~Antonova, A.~Varava, P.~Shi, J.~F. Carvalho, and D.~Kragic, ``Sequential
  topological representations for predictive models of deformable objects,'' in
  \emph{L4DC}, 2021.

\bibitem{ge2021enhancing}
Q.~Ge, T.~Richmond, B.~Zhong, T.~M. Marchitto, and E.~J. Lobaton, ``Enhancing
  the morphological segmentation of microscopic fossils through localized
  topology-aware edge detection,'' \emph{Autonomous Robots}, vol.~45, no.~5,
  pp. 709--723, 2021.

\bibitem{varava2017herding}
A.~Varava, K.~Hang, D.~Kragic, and F.~T. Pokorny, ``Herding by caging: a
  topological approach towards guiding moving agents via mobile robots.'' in
  \emph{R:SS}, 2017.

\bibitem{pokorny2016high}
F.~T. Pokorny, D.~Kragic, L.~E. Kavraki, and K.~Goldberg, ``{High-dimensional
  winding-augmented motion planning with 2D topological task projections \&
  persistent homology},'' in \emph{ICRA}, 2016.

\bibitem{carvalho2019long}
J.~F. Carvalho, M.~Vejdemo-Johansson, F.~T. Pokorny, and D.~Kragic, ``Long-term
  prediction of motion trajectories using path homology clusters,'' in
  \emph{IROS}, 2019.

\bibitem{orthey2020visualizing}
A.~Orthey and M.~Toussaint, ``Visualizing local minima in multi-robot motion
  planning using multilevel morse theory,'' in \emph{International Workshop on
  the Algorithmic Foundations of Robotics}.\hskip 1em plus 0.5em minus
  0.4em\relax Springer, 2020, pp. 502--517.

\bibitem{acar2002morse}
E.~U. Acar, H.~Choset, A.~A. Rizzi, P.~N. Atkar, and D.~Hull, ``Morse
  decompositions for coverage tasks,'' \emph{{IJRR}}, vol.~21, no.~4, pp.
  331--344, 2002.

\bibitem{morsegraph}
E.~R. Vieira, E.~Granados, A.~Sivaramakrishnan, M.~Gameiro, K.~Mischaikow, and
  K.~E. Bekris, ``{Morse Graphs: Topological Tools for Analyzing the Global
  Dynamics of Robot Controllers},'' in \emph{The 15th International Workshop on
  the Algorithmic Foundations of Robotics (WAFR)}, 2022.

\bibitem{kalies:mischaikow:vandervorst:14}
W.~D. Kalies, K.~Mischaikow, and R.~Vandervorst, ``{Lattice structures for
  attractors I},'' \emph{J. Comput. Dyn.}, vol.~1, no.~2, pp. 307--338, 2014.

\bibitem{kalies:mischaikow:vandervorst:15}
------, ``Lattice structures for attractors {II},'' \emph{{Foundations of
  Computational Mathematics}}, vol.~1, no.~2, pp. 1--41, 2015.

\bibitem{kalies:mischaikow:vandervorst:21}
------, ``{Lattice Structures for Attractors (III)},'' \emph{Journal of
  Dynamics and Differential Equations}, pp. 1572--9222, 2021.

\bibitem{CMGDB}
S.~H. Marcio~Gameiro, ``{CMGDB: Conley Morse Graph Database Software},'' 2022.

\bibitem{bogdan2022gp}
B.~Batko, M.~Gameiro, Y.~Hung, W.~Kalies, K.~Mischaikow, and E.~Vieira,
  ``Identifying nonlinear dynamics with high confidence from sparse data,''
  \emph{arXiv preprint arXiv:2206.13779}, 2022.

\bibitem{Gramacy2020}
R.~Gramacy, \emph{Surrogates: Gaussian process modeling, design and
  optimization for the applied sciences}.\hskip 1em plus 0.5em minus
  0.4em\relax Chapman Hall/CRC, Boca Raton, FL., 2020.

\bibitem{Bush:Gameiro:Harker}
J.~Bush, M.~Gameiro, S.~Harker, H.~Kokubu, K.~Mischaikow, I.~Obayashi, and
  P.~Pilarczyk, ``Combinatorial-topological framework for the analysis of
  global dynamics,'' \emph{Chaos: An Interdisciplinary Journal of Nonlinear
  Science}, vol.~22, no.~4, 2012.

\bibitem{yuan2021safe}
Z.~Yuan, A.~W. Hall, S.~Zhou, L.~Brunke, M.~Greeff, J.~Panerati, and A.~P.
  Schoellig, ``{safe-control-gym: a Unified Benchmark Suite for Safe
  Learning-based Control and Reinforcement Learning},'' \emph{arXiv preprint
  arXiv:2109.06325}, 2021.

\bibitem{moore1990efficient}
A.~W. Moore, ``Efficient memory-based learning for robot control,'' 1990.

\bibitem{meditch1964problem}
J.~Meditch, ``On the problem of optimal thrust programming for a lunar soft
  landing,'' \emph{IEEE Transactions on Automatic Control}, vol.~9, no.~4, pp.
  477--484, 1964.

\bibitem{spong_acrobot}
M.~Spong, ``{The swing up control problem for the Acrobot},'' \emph{IEEE
  Control Systems Magazine}, vol.~15, no.~1, pp. 49--55, 1995.

\bibitem{corke2011robotics}
P.~I. Corke and O.~Khatib, \emph{Robotics, vision and control: fundamental
  algorithms in MATLAB}.\hskip 1em plus 0.5em minus 0.4em\relax Springer, 2011,
  vol.~73.

\bibitem{Haarnoja2018SoftAO}
T.~Haarnoja, A.~Zhou, P.~Abbeel, and S.~Levine, ``{Soft Actor-Critic:
  Off-Policy Maximum Entropy Deep Reinforcement Learning with a Stochastic
  Actor},'' in \emph{ICML}, 2018.

\bibitem{farsi2022piecewise}
M.~Farsi, Y.~Li, Y.~Yuan, and J.~Liu, ``A piecewise learning framework for
  control of unknown nonlinear systems with stability guarantees,'' in
  \emph{Learning for Dynamics and Control Conference}.\hskip 1em plus 0.5em
  minus 0.4em\relax PMLR, 2022, pp. 830--843.

\bibitem{prajna2002introducing}
S.~Prajna, A.~Papachristodoulou, and P.~A. Parrilo, ``{Introducing SOSTOOLS: A
  general purpose sum of squares programming solver},'' in \emph{CDC}, 2002.

\bibitem{sturm1999using}
J.~F. Sturm, ``{Using SeDuMi 1.02, a MATLAB toolbox for optimization over
  symmetric cones},'' \emph{Optimization methods and software}, vol.~11, no.
  1-4, pp. 625--653, 1999.

\end{thebibliography}

\end{document}